\documentclass[letterpaper, 10 pt, conference]{ieeeconf} \IEEEoverridecommandlockouts 

\usepackage{amsmath}
\usepackage[linesnumbered, ruled, vlined]{algorithm2e}
\usepackage{graphicx}
\usepackage{eso-pic}

\IEEEoverridecommandlockouts                            
\usepackage{graphicx}
\usepackage{float}
\usepackage{atbegshi}
\usepackage{afterpage}
\usepackage{stfloats}
\usepackage{amsmath}
\usepackage{subfigure}
\usepackage{amssymb}
\usepackage{tabularx}
\usepackage{comment}

\usepackage{algpseudocode}
\usepackage{hyperref}
\usepackage{bm}
\usepackage[dvipsnames]{xcolor}
\usepackage{caption}
\usepackage{colortbl}

\usepackage[style=ieee]{biblatex}

\title{\LARGE \bf
Hierarchical Multi-Modal Planning for Fixed-Altitude Sparse Target Search and Sampling
}

\author{Lingpeng Chen, Yuchen Zheng, Apple Pui-Yi Chui, Junfeng Wu and Ziyang Hong
}

\begin{document}

\maketitle

\begin{abstract}

Efficient monitoring of sparse benthic phenomena, such as coral colonies, presents a great challenge for Autonomous Underwater Vehicles. Traditional exhaustive coverage strategies are energy-inefficient, while recent adaptive sampling approaches rely on costly vertical maneuvers. 
To address these limitations, we propose \texttt{HIMoS} (Hierarchical Informative Multi-Modal Search), a fixed-altitude framework for sparse coral search-and-sample missions. The system integrates a heterogeneous sensor suite within a two-layer planning architecture. At the strategic level, a Global Planner optimizes topological routes to maximize potential discovery. At the tactical level, a receding-horizon Local Planner leverages differentiable belief propagation to generate kinematically feasible trajectories that balance acoustic substrate exploration, visual coral search, and close-range sampling. Validated in high-fidelity simulations derived from real-world coral reef benthic surveys, our approach demonstrates superior mission efficiency compared to state-of-the-art baselines.
\vspace{-5pt}
\end{abstract}

\section{Introduction}
    \label{Sec:Intro}

    Efficient monitoring of sparse benthic phenomena, such as corals, invasive species, or rare biological aggregations, is a critical challenge for autonomous underwater vehicles (AUVs) \cite{sultan2025autonomous}. Unlike continuous scalar fields (e.g., salinity or temperature) that can be mapped using standard spatiotemporal models \cite{chen2025distributed, marchant2014bayesian}, biological targets are intrinsically discrete and sparsely distributed. Traditional exhaustive coverage strategies, such as lawnmower paths \cite{galceran2013survey, huang2021novel}, are highly energy-inefficient in these scenarios, wasting limited battery life on traversing vast, empty stretches of the seabed.

\begin{figure}[t]
    \centering
      \includegraphics[width=\linewidth]{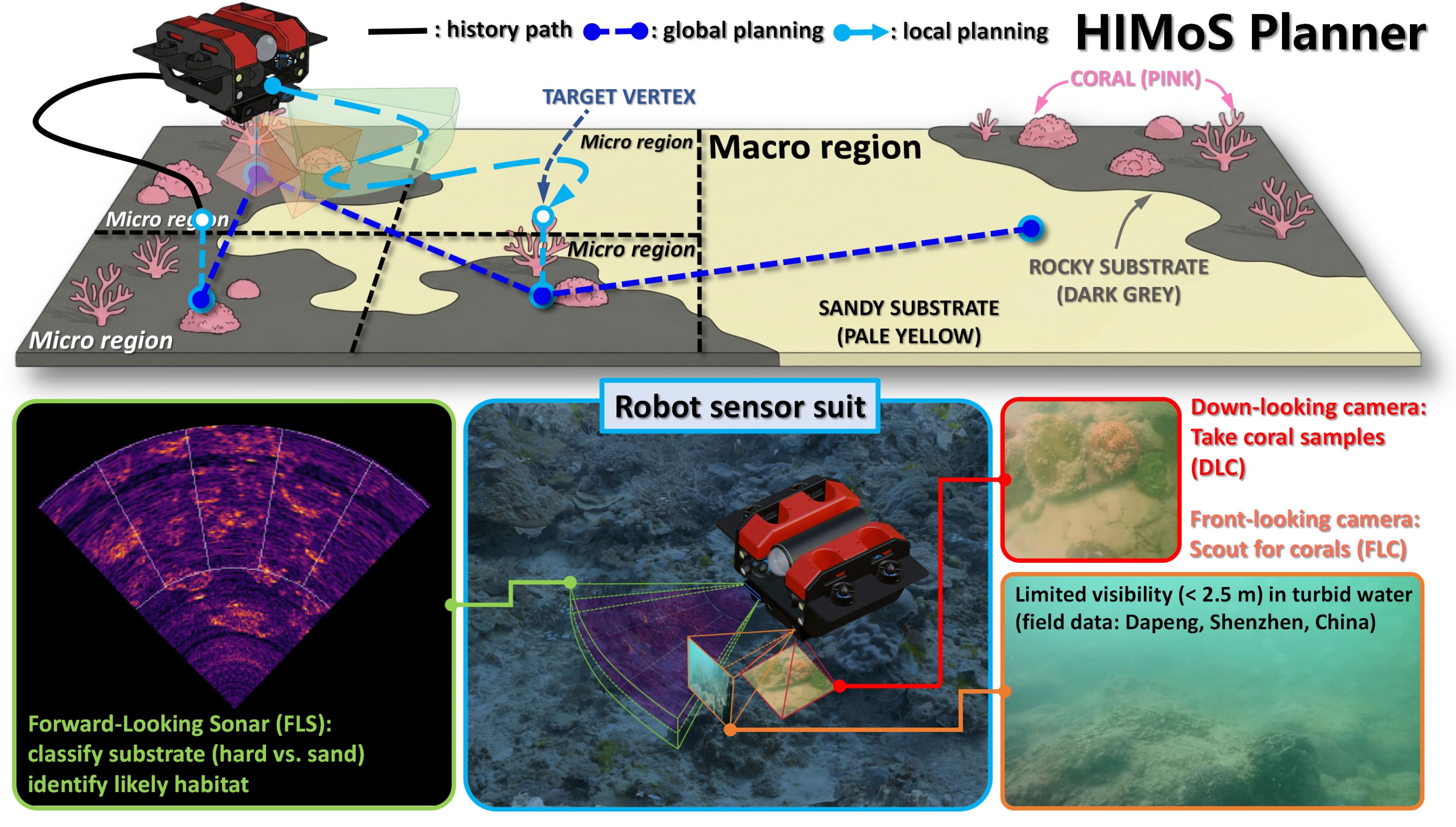}
      \caption{\texttt{HIMoS} overview. \textit{Bottom}: multi-modal sensing for turbid ocean. \textit{Top}: hierarchical planning framework steers the AUV to high-probability coral habitats for sampling.}
      \label{fig:motivation}
    \end{figure}

    Recent informative path planning and adaptive sampling methods \cite{zheng2025ergodic, meera2019obstacle} vary sensor coverage by oscillating between high-altitude visual scouting and low-altitude target sampling; this idea is exemplified by the Sparse Adaptive Search and Sample (SASS) algorithm~\cite{sass_paper}, which has been applied to coral sampling. However, these methods face significant hurdles in realistic ocean deployments. Physically, frequent vertical maneuvering is highly energy-intensive, and reliance on high-altitude vision is brittle in turbid marine environments (e.g., Shenzhen coastal waters, Fig. \ref{fig:motivation}), where light scattering severely limits optical range. Algorithmically, state-of-the-art frameworks often rely on grid-based discrete planners like Monte Carlo Tree Search \cite{sass_paper}, which scale poorly with map resolution and struggle to yield kinematically feasible trajectories. This highlights a pressing need for a fixed-altitude framework that seamlessly couples long-range robust acoustics \cite{sorensen2023commercial, chen2025sonarsweep} with close-range vision, driven by a scalable and kinematically-aware planning architecture.


    To overcome these challenges, we propose \texttt{HIMoS} (\textbf{H}ierarchical \textbf{I}nformative \textbf{M}ulti-Modal \textbf{S}earch), a robust framework tailored for sparse search-and-sample (SSS) missions. Operating safely at a fixed altitude, \texttt{HIMoS} eliminates the need for energy-intensive vertical maneuvers. Instead, it overcomes turbid marine conditions by seamlessly fusing a heterogeneous sensor suite: a Forward-Looking Sonar (FLS) provides broad-area acoustic substrate mapping \cite{garone2023seabed, xu2024seabed}, a Front-Looking Camera (FLC) enables mid-range target scouting, and a Down-Looking Camera (DLC) executes precise close-range sampling. As illustrated in Fig. \ref{fig:motivation}, this multi-modal synergy empowers the AUV to rapidly bypass vast, uninhabitable sandy regions and zero in on high-probability hard substrates where corals typically reside \cite{wright2021spatial}.
    
    Algorithmically, \texttt{HIMoS} is driven by a hierarchical architecture that seamlessly couples long-horizon strategic routing with agile local maneuvering. At the strategic level, a Global Planner identifies the most promising target regions under a strict travel budget. We cast this decision as an orienteering problem and rank candidate regions using an Upper Confidence Bound (UCB) utility that rewards both high predicted habitat probability and high epistemic uncertainty, thereby balancing exploitation and exploration.
    At the tactical level, a receding-horizon Local Planner translates these global directives into execution. To fuse heterogeneous sensors (FLS, FLC, and DLC), we introduce \textit{Differentiable Belief Dynamics} to continuously simulate the future evolution of stochastic sensor updates. By embedding these dynamics into a unified objective function, the planner generates smooth, kinematically feasible trajectories that optimally trade off multi-modal map entropy reduction (exploration) and precise visual target sampling (exploitation).

    Our main contributions are threefold. First, we propose a \textit{hierarchical global-local planning architecture} that integrates a strategic orienteering-based Global Planner with a receding-horizon Local Planner, achieving mission-level efficiency while retaining local agility. Second, we formulate local informative planning as a gradient-based trajectory optimization problem via \textit{Differentiable Belief Dynamics}. By employing a continuous surrogate for stochastic sensor updates, the planner generates kinematically feasible and non-myopic trajectories that seamlessly balance broad-area exploration with close-range target sampling. Third, we present \textit{seamless heterogeneous sensor fusion}, a unified strategy that combines acoustic and visual data to support efficient fixed-altitude sparse target search and sampling in ocean environments.
    
    
        
        
        

\section{Problem Formulation}
    We formulate the SSS mission as a multi-sensor informative path planning (IPP) problem. Consider an AUV operating in an 2D unknown ocean environment $\mathcal{M}$ and aim to maximize the number of sampled coral targets under a time budget $T_{total}$ (second). 
   
    The environment $\mathcal{M}$ is discretized into a grid map consisting of $N$ cells, indexed by $i$. Each cell possesses two latent binary states. \textbf{Substrate state} $s_i \in \{0, 1\}$ indicates habitability, where $s_i=1$ denotes hard substrate and $s_i=0$ denotes uninhabitable substrate (e.g., sand). \textbf{Target state} $c_i \in \{0, 1\}$ indicates the presence of a coral target, where $c_i=1$ denotes presence.
    
    The AUV operates at a fixed altitude suitable for benthic surveying, so the configuration space is restricted to Special Euclidean group $SE(2)$. At discrete planning step $t$, the robot state is $\mathbf{x}_t=[\mathbf{p}_t^\top,\theta_t]^\top\in SE(2)$, where $\mathbf{p}_t \in \mathbb{R}^2$ represents the position and $\theta_t$ represents the heading. The AUV system evolves according to discrete-time holonomic kinematics $\mathbf{x}_{t+1} = f(\mathbf{x}_t, \mathbf{u}_t)$ with continuous control input $\mathbf{u}_t = [v_{x,t}, v_{y,t}, \omega_t]^\top \in \mathcal{U}$, where $\mathcal{U}$ denotes the admissible control set (surge, sway, and yaw rate).

    \subsection{Heterogeneous Observation Models}
    
        As motivated in Sec. I and illustrated in Fig. \ref{fig:motivation}, the AUV is equipped with a heterogeneous sensor suite to effectively navigate turbid marine environments. To mathematically formalize their complementary roles—namely, wide-range acoustic habitat mapping (FLS) , mid-range visual target scouting (FLC), and close-range coral sampling (DLC)—we categorize the three distinct modalities into \textit{Probabilistic Scouting Sensors} and a \textit{Deterministic Verification Sensor}.

        \subsubsection{Probabilistic Scouting Sensors} 
        The FLS and FLC are tasked with gathering long-range evidence of habitats and mid-range evidence of targets, respectively. Because signal reliability inherently degrades with distance, these sensors provide noisy, range-dependent measurements over a sector-shaped Field-of-View (FOV). Let $\mathcal{V}_{\cdot}(\mathbf{x}_t, \Theta^\cdot)$ denote the set of cells visible to a sensor at robot state $\mathbf{x}_t$, where $\Theta^\cdot=\{R^\cdot_{\max},\phi^\cdot_{\mathrm{fov}}\}$ specifies the maximum sensing range $R_{\max}$ and the FOV angular span $\phi^\cdot_{\mathrm{fov}}$. For a cell $i$ centered at $\mathbf{p}_i \in \mathcal{V}_{\cdot}(\mathbf{x}_t, \Theta)$, the binary observation $z^{\cdot}_{t,i}\in\{0,1\}$ is modeled using a probabilistic detection model parameterized by the True Positive ($P_{\mathrm{TP}}$) and False Positive ($P_{\mathrm{FP}}$) rates \cite{thrun2002probabilistic}. We assume these rates are functions of the robot-to-cell distance $d_{t,i}=\|\mathbf{p}_t-\mathbf{p}_i\|$ (with $\mathbf{p}_t$ the robot position), and their profiles are known \textit{a priori} (e.g. via sensor calibration):
        \begin{equation}
            \label{eq:scouting_sensor}
            P(z^{\cdot}_{t,i} = 1 \mid \text{state}_i) = 
            \begin{cases} 
            P_{\text{TP}}(d_{t,i}) & \text{if } \text{state}_i (s_i/c_i) = 1 \\
            P_{\text{FP}}(d_{t,i}) & \text{if } \text{state}_i = 0 .
            \end{cases}
        \end{equation}

        \noindent We instantiate this general stochastic model for our two scouting modalities: the FLS  observes the substrate state $s_i$, while the FLC  observes the target state $c_i$.

        \subsubsection{Deterministic Verification Sensor}
        The DLC provides close-range verification for coral sampling with a small footprint $\mathcal{V}_{\text{DLC}}(\mathbf{x}_t)$ beneath the robot. At this proximity, we model DLC observations as noise-free and deterministic:
        \begin{equation}
        z^{\text{DLC}}_{t,j} = c_j, \quad \forall j \in \mathcal{V}_{\text{DLC}}(\mathbf{x}_t).
        \end{equation}
        Crucially, a target is counted as \textbf{sampled} (yields reward) if and only if it is observed by the DLC.

    \subsection{Environmental Prior and Belief Dynamics}
    
        The existence of a target is governed by a strong biological prior: corals exclusively inhabit hard substrates \cite{wright2021spatial}. We model this as a conditional probability constraint $ P(c_i=1 \mid s_i=0) = 0.$ Consequently, the joint probability of finding a coral is strictly coupled with the substrate probability: $P(c_i=1) = P(c_i=1 \mid s_i=1)P(s_i=1).$
        Initially, the true map layers $\{s_i\}_{i=1}^N$ and $\{c_i\}_{i=1}^N$ are unknown to the robot. To reflect this initial uncertainty, the environment is initialized with uninformative prior probabilities (e.g., uniform distributions $P(s_i=1) = 0.5$ and $P(c_i=1 \mid s_i=1) = 0.5$). Since the true states remain hidden, the robot maintains a probabilistic \textit{belief state} over $\mathcal{M}$ at each discrete time step $t$. We define this belief state as $\mathcal{B}_t = \{\mathcal{B}^S_t, \mathcal{B}^C_t\}$, where $\mathcal{B}^S_{t,i} = P(s_i=1 \mid z_{1:t}^{\text{FLS}})$ is the posterior probability of the substrate type given historical acoustic observations.
        $\mathcal{B}^C_{t,i} = P(c_i=1 \mid z_{1:t}^{\text{FLC}})$ is the posterior probability of coral existence given historical visual observations.

        These beliefs are updated via Bayesian filtering with new scouting measurements. For numerical stability, we use a log-odds form \cite{thrun2002probabilistic}. 
        Let $b^{\cdot}_{t,i}=P(\text{state}_i=1\mid z^{\cdot}_{1:t})$ and $\ell^{\cdot}_{t,i}=\log\frac{b^{\cdot}_{t,i}}{1-b^{\cdot}_{t,i}}$. 
        Based on Eq. (\ref{eq:scouting_sensor}), given a binary scouting measurement $z^{\cdot}_{t,i}\in\{0,1\}$, the log-odds update is\footnotemark 
        \begin{equation}
        \ell^{\cdot}_{t,i}=\ell^{\cdot}_{t-1,i}+ \log
        \frac{P_{\mathrm{TP}}(d_{t,i})^{z^{\cdot}_{t,i}}\!\left(1-P_{\mathrm{TP}}(d_{t,i})\right)^{1-z^{\cdot}_{t,i}}}
             {P_{\mathrm{FP}}(d_{t,i})^{z^{\cdot}_{t,i}}\!\left(1-P_{\mathrm{FP}}(d_{t,i})\right)^{1-z^{\cdot}_{t,i}}}.
        \label{eq:logodds_tpfp_compact}
        \end{equation}
        \noindent Intuitively, as the belief distributions converge and uncertainty diminishes, the robot can exploit the substrate belief map $B^S$ to find high-yield regions and transform ambiguous coral belief maps $B^C$ into confident coral sampling targets.
        
        \footnotetext{For deterministic DLC measurement (Eq.~(2)), the posterior becomes deterministic; therefore we saturate the log-odds to $[\ell^{\cdot}_{\min},\ell^{\cdot}_{\max}]$.}

    \subsection{Optimization Objective}
        We  seek a control sequence $\mathbf{U}^*=\{\mathbf{u}_0^*,\dots,\mathbf{u}_{T-1}^*\}$ that maximizes the number of targets sampled by the DLC. To avoid double-counting, we maintain a binary history mask $\xi_t\in\{0,1\}^N$ indicating cells previously covered by the DLC. The task is formulated as maximizing the cumulative reward (count of sampled corals) under the time budget $T_{total}$:
        \begin{align}
            \label{eq:objective_function}
             \mathbf{U}^* &= \operatorname*{arg\,max}_{\mathbf{u}_{0:T-1} \in \mathcal{U}^T} \mathbb{E}_{\mathcal{B}} \left[ \sum_{k=1}^{T} \sum_{j \in \mathcal{V}_{\text{DLC}}(\mathbf{x}_k)} c_j \cdot (1 - \xi_{k-1,j}) \right] \\
            \text{s.t.} \quad & \mathbf{x}_{k+1} = f(\mathbf{x}_k, \mathbf{u}_k, \Delta t), \quad T \cdot \Delta t \leq T_{total}, \nonumber
        \end{align}
        where $\mathbb{E}_{\mathcal{B}}$ is taken over the scouting beliefs ($B^S,B^C$), which steer the DLC toward high-probability targets.

\section{System Overview}
    \begin{figure*}[t]
          \centering
          \includegraphics[width=\textwidth]{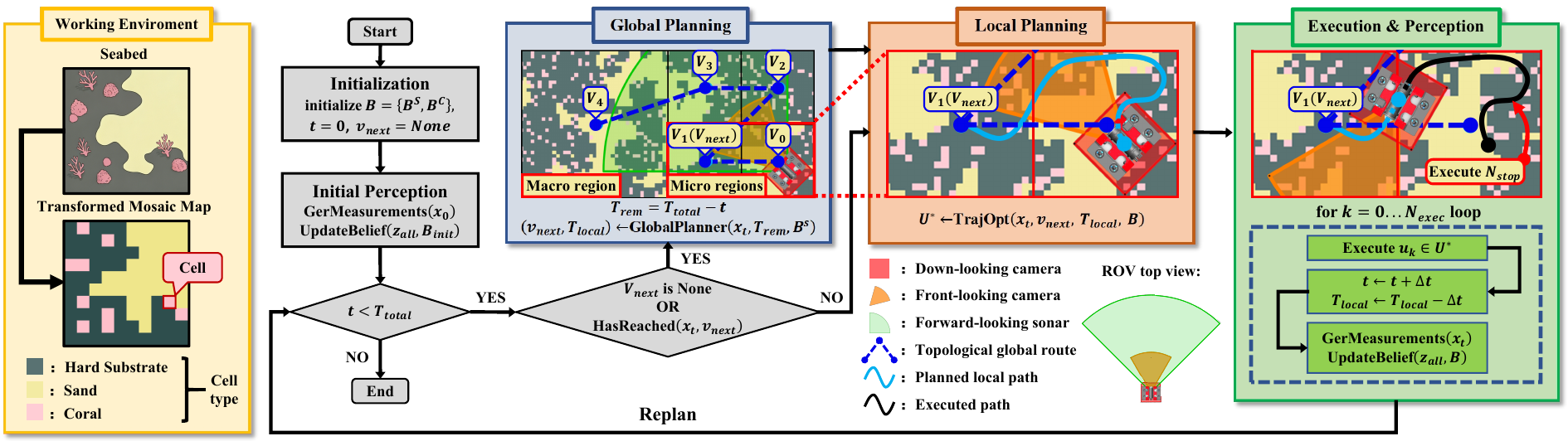} 
          \caption{System overview. An event-triggered Global Planner produces the next target region $v_{\text{next}}$ and local budget $T_{\text{local}}$. A time-triggered Local Planner receives them and optimizes a finite-horizon trajectory based on belief $\mathcal{B}$. The robot then executes the trajectory for $N_{\text{exec}}$ steps while collecting measurements and updating the belief, after which \texttt{HIMoS} re-plans.}
          \label{fig:system_overview}
          \vspace{-15pt}
        \end{figure*}

    Directly maximizing the expected sampling reward in Eq. (\ref{eq:objective_function}) is computationally intractable. To make this objective solvable while strictly satisfying $T_{total}$, we design \texttt{HIMoS} shown in Fig. \ref{fig:system_overview}, a hierarchical framework that spatially and temporally decomposes the overarching problem. 
    
    Operating in an online closed-loop manner, \texttt{HIMoS} bridges long-term strategic routing with short-term tactical execution. 
    At the strategic level, an event-triggered Global Planner ---activated whenever the AUV successfully reaches its current target region $v_{next}$ (\texttt{HasReached}) and requires new global guidance--- approximates Eq. (\ref{eq:objective_function}) by evaluating the remaining time and the latest substrate belief $\mathcal{B}^S$ to solve a topological Orienteering Problem (OP) (Sec. IV). This abstracts the global search space, generating a route across high-probability habitats. Instead of executing the entire route, the planner extracts the first OP node as the immediate target region $v_{next}$ and allocates a local time budget $T_{local}$ to constrain the subsequent tactical maneuvers.
    
    At the tactical level, a time-triggered Local Planner translates this directive into kinematically feasible maneuvers toward $v_{next}$. To bypass the non-differentiability of stochastic observations, we introduce Differentiable Belief Dynamics (Sec. V). This surrogate model casts the finite-horizon information search and sample as a smooth Non-Linear Programming (NLP) problem bounded by $T_{local}$. Operating in a receding-horizon fashion, the AUV executes only the first $N_{exec}$ steps, continuously acquiring observations to update the belief map $\mathcal{B} = \{\mathcal{B}^S, \mathcal{B}^C\}$. 
    This tightly coupled sense–update–replan loop persists locally until $v_{next}$ is reached, which triggers the next global guidance and naturally chains local executions into global milestones. This grounds every decision in the most recent perceptual evidence, enhancing overall mission efficiency and robustness against noise in turbid environments.

    \begin{figure}[b]
        \vspace{2pt}
        \centering
        \includegraphics[width=0.95\linewidth]{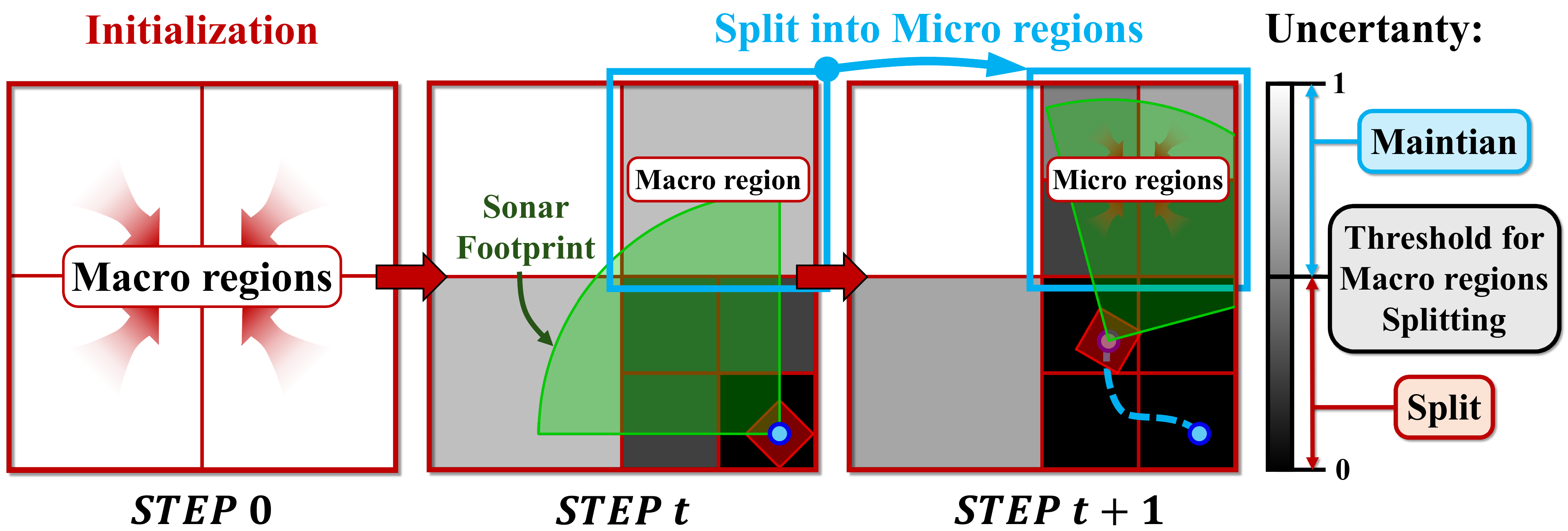}
        \caption{Adaptive Macro region splitting process. When FLS gathers enough evidence to reduce a region's substrate uncertainty below a threshold, the corresponding Macro region is divided into Micro regions to enable high-fidelity planning.}
        \label{fig:macro_micro_split}
    \end{figure}
    
\section{Global Planning: Topological Route Generation}
    \label{sec:global_planning}    

    To guide the AUV toward high-potential habitats under a strict time budget, the Global Planner formulates an Orienteering Problem (OP) over the FLS-updated substrate belief $B^S$. As outlined in Alg. 1, this process comprises three stages: (1) constructing an adaptive spatial graph from $B^S$, (2) modeling a continuous reward field via a Heteroscedastic Gaussian Process (GP), and (3) solving the OP to output the next target node $v_{next}$ and a local time budget $T_{local}$.

    \subsection{Adaptive Multi-Resolution Spatial Graph and Node Aggregation}
        \label{sec:global_graph}
        To balance decision quality and real-time tractability, we construct a multi-resolution graph $\mathcal{G} = (\mathcal{V}, \mathcal{E})$, whose node resolution is modulated by the FLS coverage encoded in $\mathcal{B}^S$.
        
        \subsubsection{Hierarchical Workspace Decomposition}
        We initialize the workspace with coarse macro regions, represented by their centroids as a node set $\mathcal{V}_{macro}$. This keeps the OP tractable during early, sparse exploration. As acoustic coverage accumulates, if the FLS observation reduces the aggregated posterior uncertainty of a macro region below a predefined threshold (Fig.~\ref{fig:macro_micro_split}), the region is subdivided into finer micro regions. Their centroids form the micro node set $\mathcal{V}_{micro}$. The resulting heterogeneous node set $\mathcal{V_{\text{opt}}} = \mathcal{V}_{macro} \cup \mathcal{V}_{micro}$ maintains broad, low-resolution nodes in unexplored areas while incorporating high-fidelity nodes for exploitation in confidently surveyed regions.
  
        \subsubsection{Probabilistic Node Aggregation}
        We extract node-level statistics from the belief map $\mathcal{B}^S_t$ for the graph-based planner (Line~6). For each micro node $v_i \in \mathcal{V}_{micro}$ representing a spatial region $\mathcal{R}_i$ centered at $\mathbf{p}_i \in \mathbb{R}^2$, 
        We aggregate the local substrate density $\bar{\rho}_i$ and its associated observation uncertainty $\bar{\nu}_i^2$ by spatially averaging the belief mean and variance over grid cells $j \in \mathcal{R}_i$, i.e., $\bar{\rho}_i = \mathbb{E}_{j \in \mathcal{R}_i}[B^S_j]$ and $\bar{\nu}_i^2 = \mathbb{E}_{j \in \mathcal{R}_i}[\mathrm{Var}(B^S_j)]$. This aggregation effectively compresses dense map beliefs into a compact graph representation $\mathcal{D} = \{(\mathbf{p}_i, \bar{\rho}_i, \bar{\nu}_i^2)\}$ to train the heteroscedastic Gaussian Process.

        \begin{algorithm}[t]
            \linespread{1.1}\selectfont
            \DontPrintSemicolon
            \caption{Global Planner}
            \label{alg:global_planner}
            
            \SetKwFunction{Init}{InitMacroNodes}
            \SetKwFunction{IsObserved}{IsBlockObserved}
            \SetKwFunction{Split}{SplitMacroBlock}
            \SetKwFunction{Aggregate}{AggregateNodeStats}
            \SetKwFunction{TrainGP}{TrainGP}
            \SetKwFunction{CalcRew}{CalcInfoReward}
            \SetKwFunction{SolveOP}{SolveOP}
            
            \SetKwInOut{Input}{Input}
            \SetKwInOut{Output}{Output}
            \SetKwInOut{State}{State}
        
            \Input{State $\mathbf{x}_t$, Substrate Belief $\mathcal{B}^S$, Time $T_{rem}$}
            \Output{Target $v_{next}$, Local Budget $T_{local}$}
            \State{$\mathcal{V}_{macro} \leftarrow \text{Init}(\mathcal{M})$, $\mathcal{V}_{micro} \leftarrow \emptyset$}
         
            \smallskip
            \vspace{-5pt}
            \tcp{-- 1. Construct graph --}
            \ForEach{$v \in \mathcal{V}_{macro}$}{
                \If{\IsObserved{$v, \mathcal{B}^S_t$}}{
                    $\mathcal{V}_{macro} \leftarrow \mathcal{V}_{macro} \setminus \{v\}$\;
                    $\mathcal{V}_{micro} \leftarrow \mathcal{V}_{micro} \cup$ \Split{$v$}\;
                }}
            \vspace{-3pt}
            $\mathcal{V}_{opt} \leftarrow \mathcal{V}_{macro} \cup \mathcal{V}_{micro}$ 
            
            \tcp{-- 2. Learn Reward Field --}
        
            $\mathcal{D} \leftarrow$ \Aggregate{$\mathcal{V}_{micro}, \mathcal{B}^S_t$}\;
            
            $GP \leftarrow$ \TrainGP{$\mathcal{D}$}\;
            
            \ForEach{$v_i \in \mathcal{V}_{opt}$}{
                $r_i \leftarrow$ \CalcRew{$GP, v_i$} \;
            }
            
            \tcp{-- 3. Solve OP --}
            $D_{rem} \leftarrow T_{rem} / c$ \tcp*[r]{\scriptsize Map time to distance budget} 
            $\pi^* \leftarrow$ \SolveOP{$\mathcal{V}_{opt}, \{r_i\}, D_{rem}$}\;
            
            $v_{next} \leftarrow \pi^*.pop()$ \quad $e_{next} \leftarrow \text{Dist}(\mathbf{x}_t, v_{next})$\;
            $T_{local} \leftarrow c \cdot \|e_{next}\|$ \tcp*[r]{\scriptsize Map back to tactical time}
        
            \Return{$v_{next}, T_{local}$}
        \end{algorithm}

    \subsection{Heteroscedastic Substrate Modeling}

        To guide the robot toward promising habitats, we model the latent substrate density field as a continuous function. Because the reliability of acoustic measurements inherently degrades with sensing range, we employ a heteroscedastic Gaussian Process (GP) to rigorously account for spatially varying observation noise~\cite{rasmussen2003gaussian}. Trained on the aggregated graph statistics $\mathcal{D}$, the GP constructs an observation noise matrix $\mathbf{R} = \text{diag}(\bar{\nu}_1^2, \dots, \bar{\nu}_M^2)$ that explicitly injects each node's input-dependent uncertainty. For any candidate node $v_i$ at $\mathbf{p}_i$, the posterior predictive mean and variance are computed directly as $\mu_i = \mathbf{k}_i^\top (\mathbf{K} + \mathbf{R})^{-1} \boldsymbol{\bar{\rho}}$ and $\sigma_i^2 = k(\mathbf{p}_i, \mathbf{p}_i) - \mathbf{k}_i^\top (\mathbf{K} + \mathbf{R})^{-1} \mathbf{k}_i$, respectively, where $\mathbf{K}$ is the training covariance matrix, $\mathbf{k}_i$ is the cross-covariance vector, and $\boldsymbol{\bar{\rho}} =[\bar{\rho}_1, \dots, \bar{\rho}_M]^\top$. Consequently, sparsely observed regions naturally exhibit high predictive variance $\sigma_i^2$ to drive exploration, whereas thoroughly scanned regions are ``pinned'' with low variance for confident exploitation.

    \subsection{Decision Making: The Orienteering Problem}

        At each global replanning step $t$, we determine the next strategic waypoint $v_{next}$ on the adaptive graph $\mathcal{G}_t = (\mathcal{V}_t, \mathcal{E}_t)$ by formulating a Budgeted Orienteering Problem (OP)~\cite{vansteenwegen2011orienteering}. To balance the exploitation of high substrate density regions against the exploration of unknown regions, we assign a composite reward $r_i$ to each candidate node $v_i$:
        \begin{equation}
            r_i = (\mu_i + \beta \sigma_i) \cdot \lambda_i^{area} \cdot \mathbb{I}(v_i \notin \mathcal{V}_{visit}).
        \end{equation}
        The base score follows an Upper Confidence Bound (UCB) formulation based on the GP posterior moments $(\mu_i, \sigma_i^2)$, where the exploration weight $\beta$ controls the trade-off between exploitation (high mean) and exploration (high uncertainty). The score is modulated by two heuristic factors: $\lambda_i^{area}$ scales the reward proportionally to the node's spatial area (ensuring fairness between macro and micro nodes), and the indicator function $\mathbb{I}(\cdot)$ acts as a binary filter that zeroes the reward of visited nodes.
        
        The OP seeks a path $\pi^*$ that maximizes the cumulative reward $\sum_{v_j \in \pi^*} r_j$ under a mission budget constraint. 
        To map the temporal mission budget onto the spatial graph, we convert the remaining time $T_{rem}$ into a distance budget $D_{rem} = T_{rem} / c$. 
        Here, $c$ acts as an amortized time-allocation factor that inherently inflates kinematic travel estimates, buffering sufficient margins for robot inter-region transits and intra-region tactical sampling. We solve this constrained OP efficiently using an Iterated Local Search heuristic initialized via randomized greedy selection \cite{feo1995greedy}.

        Operating in a receding-horizon manner, the Global Planner extracts only the first target node $v_{next} \in \pi^*$ and its corresponding edge $e_{next}$. This edge dictates a tactical Local Time Budget $T_{local} = c \cdot \|e_{next}\|$. By passing  the directive tuple $(v_{next}, T_{local})$ to the Local Planner, tactical maneuvers are bounded by strategic constraints, providing a stable execution horizon for information-rich maneuvers while preventing mission overtime.
        Global replanning is triggered when the robot reaches the region of the current target node $v_{next}$, decoupling high-level route planning from the high-frequency local control loop.

\section{Local Planning: Unified Receding Horizon Information Search}
    \label{sec:local_planning_unified}

    Given the global directives ($v_{next},T_{local}$), the Local Planner generates a kinematically feasible, non-myopic trajectory towards $v_{next}$. It balances two objectives: (1) exploration: reducing map entropy via FLS/FLC scouting over both the substrate map $\mathcal{B^S}$ and coral map $\mathcal{B^C}$, and (2) exploitation: precise DLC verification and targets sampling. To formulate this multi-sensor task as an efficient gradient-based Non-Linear Programming (NLP) problem, we propose a \textit{Differentiable Belief Dynamics} framework. This deterministic surrogate models the expected evidence gain from stochastic observations as a continuous process, yielding smooth gradients that enable direct optimization of the robot's control sequence to maximize information gain.

    \subsection{Multi-Modal Belief Snapshot \& Candidate Identification}

        At each planning cycle $t$, the Local Planner initializes the optimization horizon by capturing a snapshot of the latest posterior beliefs $\mathcal{B}^S_t$ and $\mathcal{B}^C_t$ (updated via Eq. 3). These snapshot log-odds, $\ell^S_t$ and $\ell^C_t$, serve as priors for the differentiable belief dynamics, thereby decoupling the evaluation of future information gain within the planning rollout from the online-evolving knowledge state. Concurrently, to direct the DLC's close-range sampling, we extract a discrete candidate map $\mathcal{M}_{samp} = \{ m_i^{\mathrm{samp}} \}_{i=1}^{N}$ from the coral belief $\mathcal{B}^C_t$. A cell $i$ is flagged as a potential sampling target if its existence probability $P(c_i=1)$ exceeds a confidence threshold $\delta$:
        \begin{equation*}
            m^{\text{samp}}_i =  \mathbb{I}(P(c_i=1)> \delta) = \mathbb{I}(\sigma(\ell^C_{t,i}) > \delta), \quad \forall i \in \mathcal{M},
        \end{equation*}
        where $\sigma(\cdot)$ denotes the logistic sigmoid mapping log-odds to probability.
        \vspace{-2pt}

    \subsection{Differentiable Belief Dynamics}
        
        The primary challenge in trajectory optimization is the incompatible nature of the sensor models: discrete FOV boundaries lack the smooth gradients necessary for NLP, while stochastic scouting (FLS/FLC) yields intractable entropy evolution. 
        To resolve this, we formulate a deterministic surrogate model. Instead of predicting the exact stochastic posterior, it tracks expected accumulation of observation evidence, yielding continuous, differentiable information states.
        
        For scouting modalities, the log-odds belief evolves as a random walk. Because active exploration fundamentally aims to push beliefs away from ambiguity ($l \approx 0$) regardless of the measurement outcome, we model the expected magnitude of this update. We define the \textit{Accumulated Observation Confidence} $\Lambda \in[0, \infty)$ as a proxy state, initialized with the absolute log-odds snapshots: $\Lambda_{0,i}^{FLS} = |\ell_{t,i}^S|$ and $\Lambda_{0,i}^{FLC} = |\ell_{t,i}^C|$. Over the prediction horizon, $\Lambda$ accumulates the sensor's \textit{Average Evidence Magnitude} $\eta(d)$, which quantifies the expected log-odds increment per observation (effectively the Signal-to-Noise Ratio at observing distance $d$):
        \begin{equation}
            \eta(d) = \frac{1}{2} \left( \left| \log \frac{P_{\text{TP}}(d)}{P_{\text{FP}}(d)} \right| + \left| \log \frac{1 - P_{\text{FP}}(d)}{1 - P_{\text{TP}}(d)} \right| \right). 
        \end{equation}
        Therefore, the deterministic dynamics for both substrate (via FLS) and target beliefs (via FLC) can be formulated as:
        \begin{equation}
            \Lambda_{k+1,i}^{\text{sensor}} = \Lambda_{k,i}^{\text{sensor}} + \eta_{\text{sensor}}(d_{k,i}) \cdot \alpha_i^{\text{sensor}}(\mathbf{x}_k), 
            \label{eq:lambda_accum}
        \end{equation}
        where $\alpha_i^{\text{sensor}}$ is a differentiable sensor field proxy that approximates the traditional binary observation indicator (i.e., 1 if observed, 0 otherwise), detailed in Sec. V-C). By maximizing $\Lambda$, the optimizer  naturally drives the vehicle toward vantage points that yield high-SNR observations.

        To integrate this accumulated evidence into the optimization cost, we map $\Lambda$ to a proxy entropy $H_{\text{proxy}}$. Since $\Lambda$ represents an optimistic upper bound on true stochastic confidence (via the triangle inequality $|l + \Delta l| \leq |l| + |\Delta l|$), it yields a lower-bound approximation of future uncertainty. We define $H_{\text{proxy}}(\Lambda)$ by converting $\Lambda$ to a pseudo-probability $\hat{p} = \sigma(\Lambda)$ and applying the standard binary entropy formula:
        \begin{equation}
            H_{\text{proxy}}(\Lambda) = \ln(1+e^\Lambda) - \Lambda \cdot \sigma(\Lambda).
        \end{equation}
        This convex and monotonically decreasing surrogate provides consistent gradients to drive uncertainty reduction.
        
        Conversely, for flagged candidate cells ($m^{\text{samp}}_i = 1$), we track a \textit{Cumulative Verification Intensity} $\Lambda^{\text{down}}$ —-a proxy state representing the accumulated DLC coverage over the target. Its dynamics mirror Eq. \ref{eq:lambda_accum} but utilize a constant, high information rate $\eta_{\text{down}}$ for the DLC. We map $\Lambda^{\text{down}}$ to a sampling probability via a saturation function:
        \begin{equation}
            P^{\text{samp}}(\Lambda_{k,i}^{\text{down}}) = 1 - \exp(-\lambda \cdot \Lambda_{k,i}^{\text{down}}).
        \end{equation}
        This formulation enforces diminishing returns: as $\Lambda^{\text{down}}$ grows, the vanishing gradient naturally prevents the robot from loitering over a sampled target, incentivizing efficient progression to subsequent candidates.
        
    \subsection{Differentiable Sensor Field Proxy}
      \begin{figure}
            \centering
            \includegraphics[width=\linewidth]{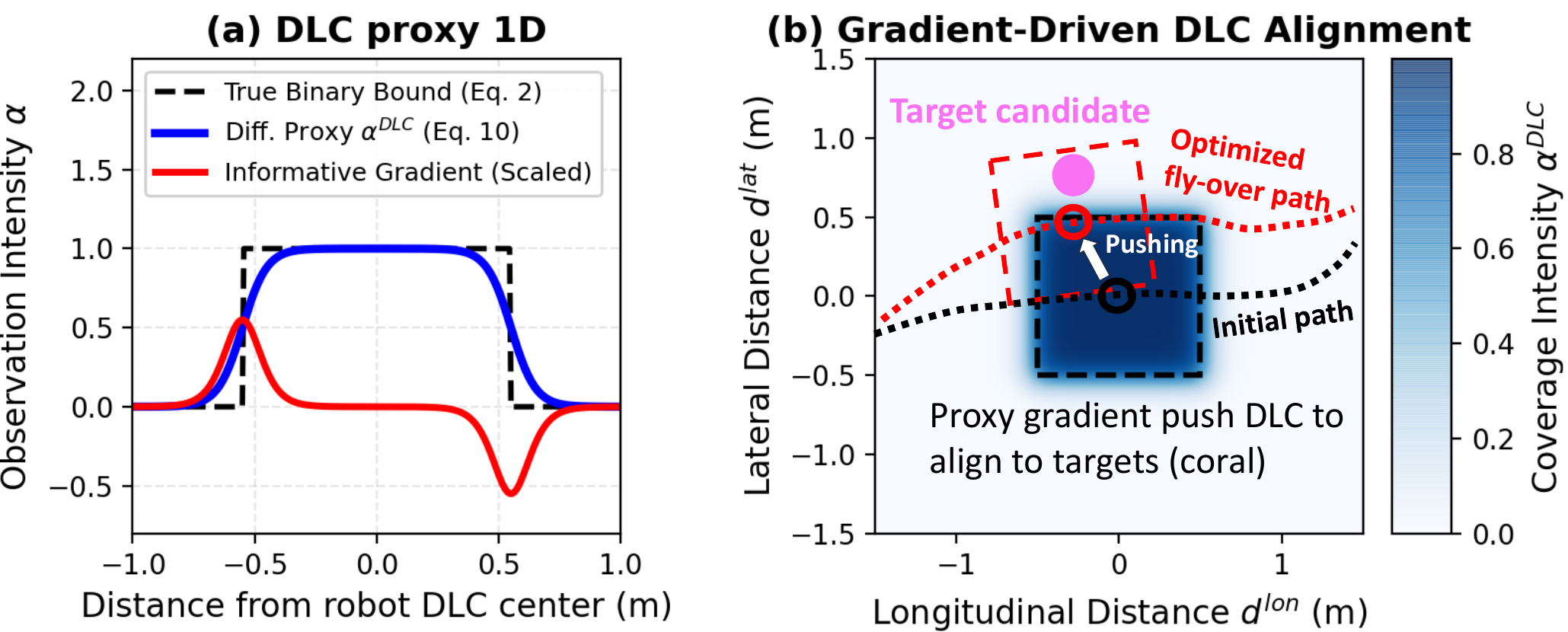}
            \caption{Differentiable DLC proxy. (a) Soft visibility indicator $\alpha^{DLC}$ and gradient near the footprint boundary. (b) The resulting gradient steers the trajectory to align the footprint with target candidates.}
            \label{fig:proxy}
        \end{figure}

        To enable gradient-based optimization, the  soft visibility indicator at cell $i$, $\alpha^\cdot_i(\mathbf{x}_t) \in [0, 1]$, of each sensor is modeled as a differentiable function of the robot state $\mathbf{x}_t=[\mathbf{p}_t^\top,\theta_t]^\top$ and cell coordinate $\mathbf{p}_i$. For FLS and FLC, we approximate their sector-shaped FOV (defined by maximum range $R_{\text{max}}$ and half-angle $\beta$) using the product of sigmoids. Let $\Delta \mathbf{p}_{t,i} = \mathbf{p}_i - \mathbf{p}_t$ and $\phi_{t,i}$ be the relative bearing to the cell.
        \begin{equation*}
            \alpha^\cdot_i(\mathbf{x}_t) = \sigma \big( \gamma_d(R_{\text{max}} - \|\Delta \mathbf{p}_{t,i}\|) \big) \cdot \sigma \big( \gamma_a(\cos \phi_{t,i} - \cos \beta) \big),
        \end{equation*}
        where $\gamma_d, \gamma_a$ control the boundary steepness. This yields two specific instantiated models: $\alpha^{FLS}$ and $\alpha^{FLC}$.
                
        Similarly, for the DLC with a square footprint of side $L_{DLC}$, we project the relative position $\Delta \mathbf{p}_{t,i}$ into the robot's body frame via the 2D rotation matrix $\mathbf{R}(\theta_t)$ to obtain the signed forward (longitudinal) and sideways (lateral) body-frame components: $[d^{\text{lon}}_{t,i}, d^{\text{lat}}_{t,i}]^\top = \mathbf{R}^\top(\theta_t)\Delta \mathbf{p}_{t,i}$. The soft footprint mask then smoothly approximates the hard square boundaries ($|d| \leq L_{DLC}/2$) via:
        \begin{equation}
            \alpha^{DLC}_i(\mathbf{x}_t) = \prod_{d \in \{d^{\text{lon}}_{t,i}, d^{\text{lat}}_{t,i}\}} \sigma \left( \frac{L_{DLC}/2 - |d|}{\epsilon} \right).
        \end{equation}
       Unlike hard binary FOV boundaries that yield zero gradients outside the footprint, the proposed smooth proxy maintains a non-vanishing gradient in the vicinity of the boundary (Fig. \ref{fig:proxy}a). During optimization, this gradient provides a consistent descent direction that pulls the vehicle state—both position and heading—toward configurations that better align the DLC footprint with target candidates (Fig. \ref{fig:proxy}b).

    \subsection{Unified Optimization Formulation}
            
        We formulate the finite-horizon trajectory optimization as a monolithic NLP problem. Given the temporal budget $T_{\text{local}}$ from the Global Planner, we define the horizon $H = \lfloor T_{\text{local}}/\Delta t \rfloor$. The Local Planner solves for an optimal finite-horizon control sequence $\mathbf{U}_{local}^*=\{\mathbf{u}_0,\ldots,\mathbf{u}_{H-1}\}$, which induces the corresponding trajectory $\tau^*$ to minimize a unified cost function $J(\tau) = J_{\text{scout}} + J_{\text{samp}} + J_{\text{reg}}$, effectively trading off acoustic exploration, visual search, and target sampling against trajectory smoothness and control effort.
                
        To drive multi-modal exploration, $J_{\text{scout}}$ penalizes the negative expected entropy reduction for both the Substrate ($\mathcal{B}^S$) and Target beliefs ($\mathcal{B}^C$). For computational tractability, we evaluate this over a downsampled active map $\mathcal{M}_{\text{act}}$ (obtained by pooling to form coarser map with the same spatial extent):
        \begin{equation*}
            J_{\text{scout}} = - \sum_{k=0}^{H-1} \sum_{i \in \mathcal{M}_{\text{act}}} \left[ w_{\text{s}} \Delta H(\Lambda_{k,i}^{\text{FLS}}) + w_{\text{c}} \Delta H(\Lambda_{k,i}^{\text{FLC}}) \right],
        \end{equation*}
        where $\Delta H(\Lambda_{k,i}) = H_{\text{proxy}}(\Lambda_{k,i}) - H_{\text{proxy}}(\Lambda_{k+1,i})$. To ensure exploitation, $J_{\text{samp}}$ drives DLC fly-overs by maximizing the expected sampling success over candidate map $\mathcal{M}_{\text{samp}}$:
        \begin{equation*}
            J_{\text{samp}} = -w_{\text{samp}} \sum_{i \in \mathcal{M_{\text{samp}}}} m_i^{\text{samp}} \cdot P^{\text{samp}}(\Lambda_{H,i}^{\text{down}}).
        \end{equation*}
                
        Finally, $J_{\text{reg}}$ enforces kinematic feasibility and smoothness via quadratic penalties on control energy and jerk, alongside a terminal cost to guide the robot toward the global waypoint $v_{next}$. The NLP is subject to system kinematics $\mathbf{x}_{k+1} = f(\mathbf{x}_k, \mathbf{u}_k)$, belief dynamics (Sec. V-B), and actuation limits. We discretize the problem via Direct Multiple Shooting and solve it using \textit{CasADi} and \textit{IPOPT} \cite{wachter2006implementation, andersson2019casadi}, which efficiently exploits the sparsity of the belief dynamics to guarantee real-time, information-rich trajectory generation.

\section{Experiments and Results}    

\begin{figure}[h]
      \includegraphics[width=\linewidth]{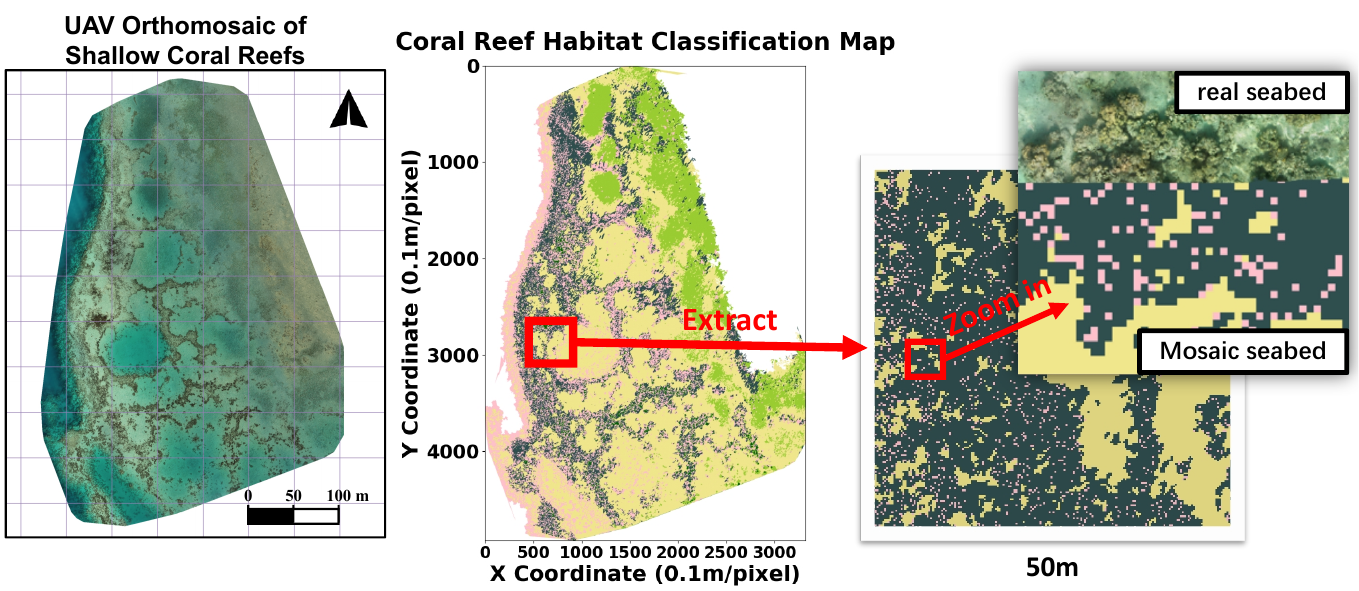} 
      \caption{Dataset-to-simulator pipeline. }
    \label{fig:dataset_generation}
    \vspace{-5pt}
    \end{figure}

    We evaluate \texttt{HIMoS} in a high-fidelity simulation environment derived from real-world UAV-based photogrammetry of shallow coral reefs shown in Fig.~\ref{fig:dataset_generation}~\cite{nieuwenhuis2022integrating}. We benchmark our approach against three baselines (key hyperparameters are summarized in Tab.~\ref{tab:exp_settings}):
    \texttt{Boustrophedon} is a standard lawnmower-style exhaustive coverage strategy. 
    \texttt{MCTS} represents the core non-myopic planning framework used in state-of-the-art coral sampling algorithms (e.g., SASS \cite{sass_paper}), which we port to our fixed-altitude, multi-sensor setting to ensure a fair comparison.
    We also provide an upper-bound baseline  \texttt{With Prior} that assumes access to the ground-truth substrate density map.  It computes a global Orienteering Problem (OP) route offline using standard solvers, and executes using the same receding-horizon Local Planner as \texttt{HIMoS}. 
    
    \begin{table}[h]
        \centering
        \scriptsize
        \caption{Key experiment settings}
        \label{tab:exp_settings}
        \setlength{\tabcolsep}{4pt}
       
        \begin{tabularx}{\linewidth}{l X}
        \hline
        \textbf{Robot} & $v_{\max}=0.5$ m/s, $\omega_{\max}=1.0$ rad/s \\
        \textbf{Planning} & Frequency $2$ Hz,  $\Delta t = 0.5$ sec \\
        \textbf{\texttt{MCTS} action space} & 
        $\{[v,0,0], [0,\pm v,0], [v_d,\pm v_d,0],[v,0,\pm \omega/2],[v,0,\pm \omega]\}$ \newline
        $v=0.5$ m/s, $\omega=1.0$ rad/s, $v_d=v/\sqrt{2}$ \\
        
        \hline
        \textbf{\texttt{HIMoS}}  & \\
        \textit{Global planner} & Macro region $=4$\,m, Micro region $=2$\,m \newline Empirical time-allocation factor $c = 6$ \newline
        UCB exploration weight $\beta = 0.6$ \\
        \textit{Local planner}: & Weights $[w_{sub}, w_{search}, w_{samp}] = [1,10,5]$ \newline
        Candidate threshold $ \delta = 0.8$, \newline
        execution steps $N_{exec}=4$ \\

        \hline
        \textbf{Sensor update} & Frequency $2$ Hz (two effective measurements per second) \\
        \textbf{FLS (substrate)} & Range $R=6.0$ m, FOV $90^\circ$;\newline $P_{\mathrm{TP}}=1-0.1d_{\mathrm{norm}},\ P_{\mathrm{FP}}=0.1d_{\mathrm{norm}}$ \\
        \textbf{FLC (target)} & Range $R=2.5$ m, FOV $60^\circ$;\newline $P_{\mathrm{TP}}=1-0.15d_{\mathrm{norm}},\ P_{\mathrm{FP}}=0.15d_{\mathrm{norm}}$ \\
        \textbf{DLC (verify)} & Footprint $1.0 \times 1.0$ m (square) \\
        \hline
        \end{tabularx}
        
        \footnotesize\emph{Note:} $d_{\mathrm{norm}}=d/R$ is the normalized distance, where $d$ is the distance to the target and $R$ is the corresponding sensor range.
    \end{table}

    \subsection{Simulation Environment and Dataset}
        We utilize the Object-Based Image Analysis (OBIA) semantic segmentation maps from \cite{nieuwenhuis2022integrating} as the simulation ground truth, inherently preserving the natural spatial clustering of benthic habitats. To create a planning-ready representation, the original semantic labels are rasterized into $0.25$\,m $\times 0.25$\,m grid cells. We consolidate the diverse benthic classes into three task-relevant categories: \textit{sand}, \textit{hard substrate}, and \textit{coral}. Because the original annotations provide habitat-level boundaries rather than discrete target locations, we instantiate individual coral targets by probabilistically sampling cells within the coral-labeled regions. This process yields discrete, sparsely distributed targets while rigorously maintaining the authentic spatial layout of the seabed.

\begin{figure}[tbp]
        \centering
          \includegraphics[width=0.9\linewidth]{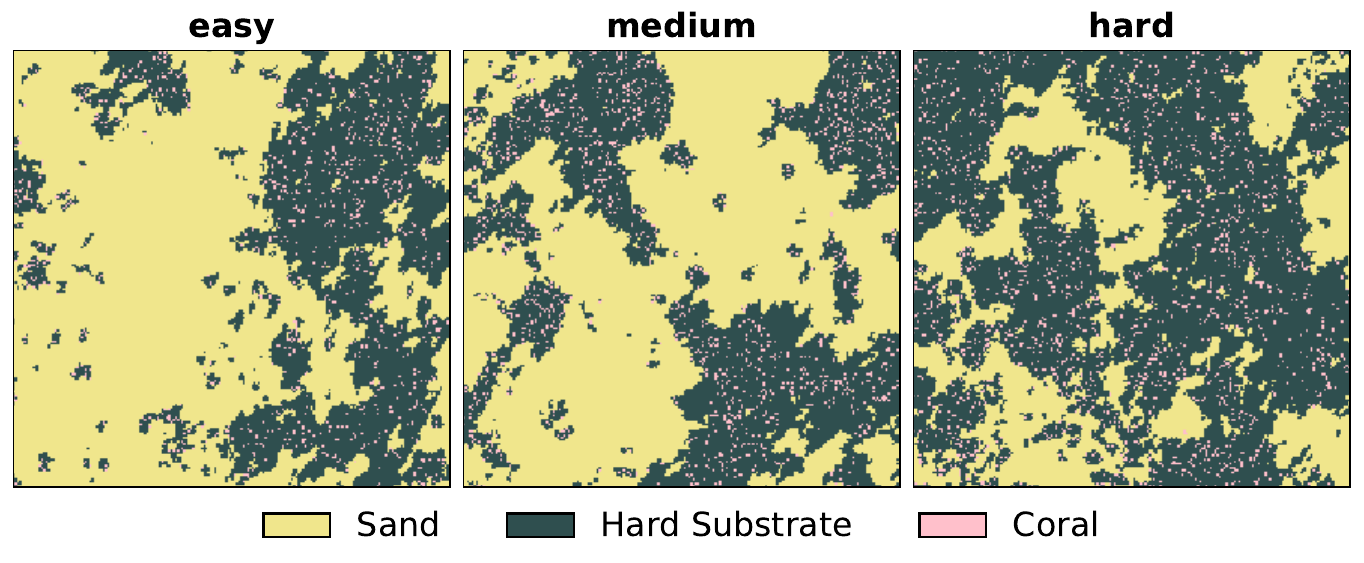} 
          \caption{Representative test maps ($50\,\mathrm{m} \times 50\,\mathrm{m}$).}
        \label{fig:dataset_examples}
        \end{figure}

        For systematic benchmarking, we extract 12 distinct $50$\,m $\times 50$\,m sub-maps from the global survey area. As shown in Fig.~\ref{fig:dataset_examples}, these maps exhibit substantial variation in coral prevalence and substrate clustering. To thoroughly evaluate algorithmic robustness, we categorize these sub-maps into \textit{easy}, \textit{medium}, and \textit{hard} subsets based on target counts and the spatial fragmentation of the hard substrates.

        \begin{figure*}[htbp]
        \centering
        \includegraphics[width=\textwidth]{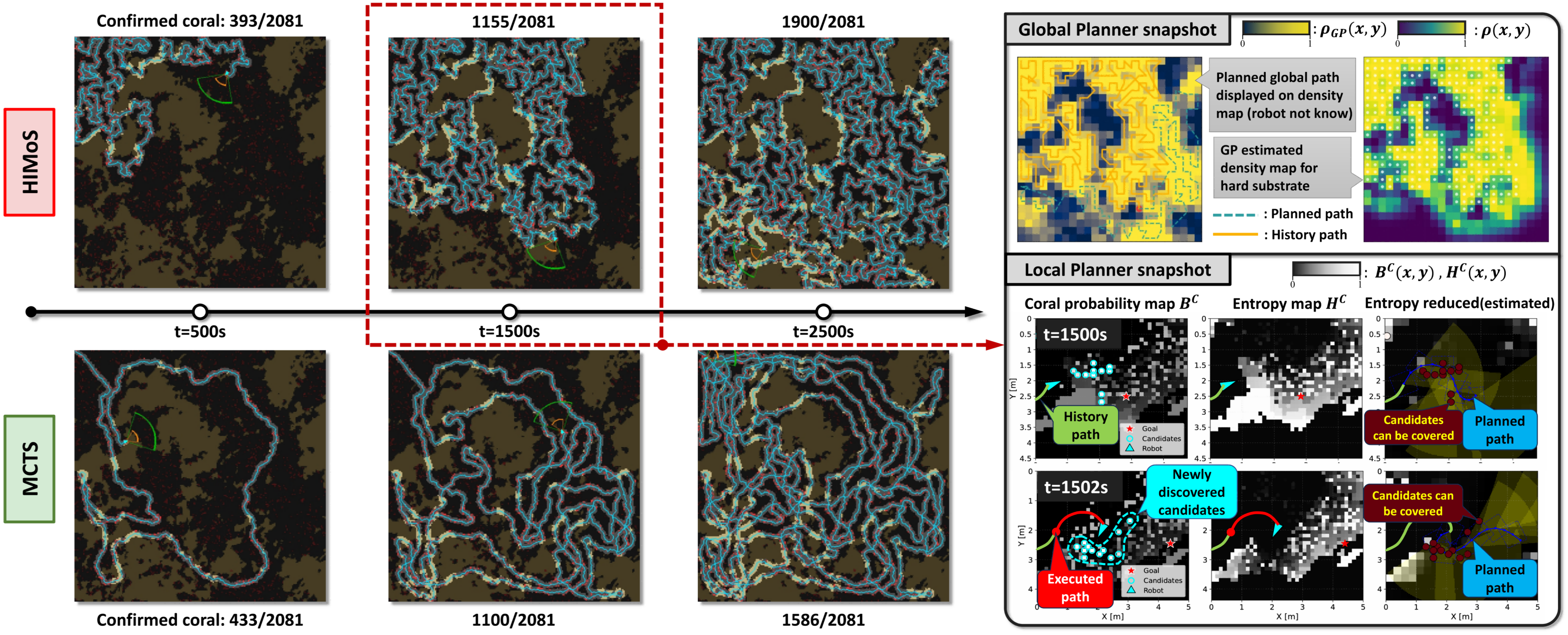}         
        \caption{Qualitative trajectory analysis.Left: trajectory evolution for \texttt{HIMoS} and \texttt{MCTS} at $500/1500/2500\,\text{s}$. Right: \texttt{HIMoS} planning snapshots at $1500\,\text{s}$, showing the Global Planner (top) and Local Planner (bottom).}
        \label{fig:trajectory_analysis}
        \vspace{-10pt}
    \end{figure*}

    \subsection{Performance Analysis}

\begin{figure}[tbp]
            \centering
            \includegraphics[width=\linewidth]{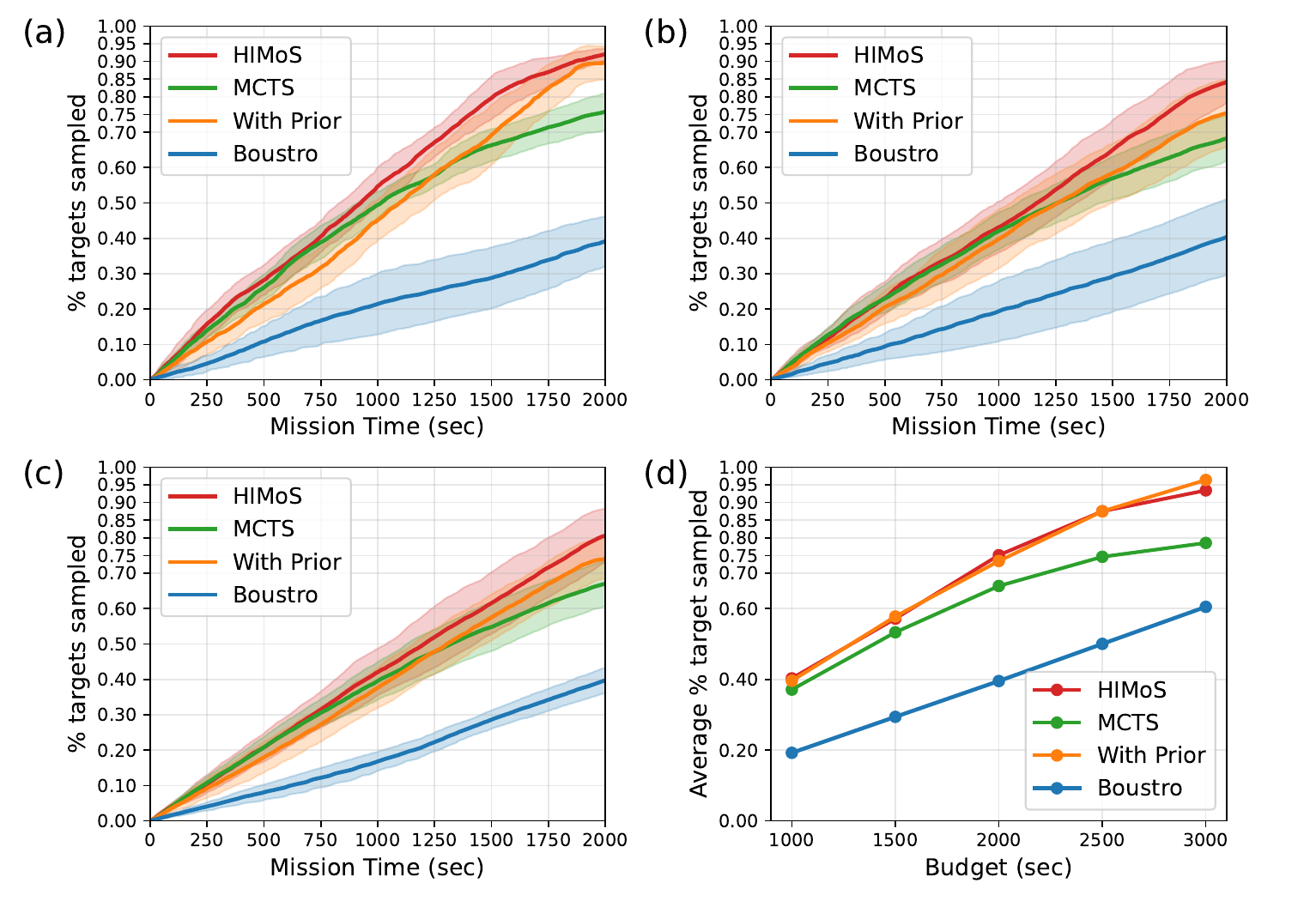}
            \caption{(a-c) Sampling ratio vs. mission time under a fixed $2000\,\mathrm{s}$ budget across \emph{easy}, \emph{medium}, and \emph{hard} environments. Solid curves denote the mean, with shaded areas indicating standard deviation. (d) Average final ratio across varying time budget in the two hardest scenarios}
            \label{fig:performance_curves} 
        \end{figure}

        To systematically evaluate algorithmic robustness, we conduct 48 independent runs per baseline by varying starting locations across the 12 maps. Fig.~\ref{fig:performance_curves}(a-c) presents the target confirmation ratio (percentage of sampled targets) over a $2000$\,s budget. \texttt{HIMoS} consistently outperforms both \texttt{MCTS} and \texttt{Boustrophedon} across all difficulty levels. Remarkably, \texttt{HIMoS} achieves a marginally higher confirmation rate than the \textit{With Prior} baseline, despite lacking initial environmental knowledge. This counter-intuitive result highlights a fundamental limitation of offline planning: solving the OP over an entire dense map suffers from the curse of dimensionality, yielding sub-optimal open-loop routes even given generous computation time ($100$\,s). In contrast, \texttt{HIMoS} dynamically decomposes the massive OP into tractable sub-problems via its adaptive multi-resolution graph, continuously refining the global strategy with recent evidence.
        
        Also, the temporal curves reveal diminishing marginal utility in the baselines. While \texttt{MCTS} is competitive early on, its discovery rate plateaus after $1500$\,s. Evaluating performance under varying mission budgets ($1000$\,s to $3000$\,s) on the two hardest maps (Fig.~\ref{fig:performance_curves}d) further illustrates this. \texttt{MCTS} shows marginal gains after $2000$\,s, whereas \texttt{HIMoS} maintains a high and steady confirmation rate across different budgets. The underlying kinematic behaviors and planning logic driving this performance gap are detailed in Sec~\ref{sec:trajectory_analysis}.

    \subsection{Trajectory Behavior Analysis}
        \label{sec:trajectory_analysis} 
        
        To illustrate the underlying planning behaviors, we visualize a $2500$\,s mission on a challenging map. Fig.~\ref{fig:trajectory_analysis} (left) compares the trajectory evolution. Initially at $500$\,s, \texttt{MCTS} explores aggressively, yielding a higher sample count than \texttt{HIMoS} (433 vs. 393). 
        However, under a strict $0.5$\,s rollout limit to maintain $2$Hz planning, the large branching factor causes the search tree to grow exponentially with depth, which in turn truncates the practical look-ahead horizon.
        Lacking global structure, \texttt{MCTS} struggles to identify distant unvisited habitats once a local region is depleted. Consequently, by $2500$\,s, it wastes energy traversing previously swept areas, stalling at a $76\%$ confirmation rate.
        
        Conversely, \texttt{HIMoS} exhibits deliberate, region-by-region exploration and exploitation, ultimately achieving a $91\%$ confirmation rate. As shown in the Global Planner snapshot at $1500$\,s (Fig.~\ref{fig:trajectory_analysis}, top right), \texttt{HIMoS} continuously regresses acoustic observations into a substrate density map. This enables a strategic, long-horizon topological skeleton that efficiently bridges unvisited high-probability rocky habitats, fundamentally preventing redundant traversals.
        
        Guided by the latest global waypoint (red star), the Local Planner orchestrates tactical maneuvers (Fig.~\ref{fig:trajectory_analysis}, lower right)\footnotemark \footnotetext{The FLS sector and substrate belief and entropy map are omitted for visual clarity.}. Driven by the differentiable belief dynamics, the solver generates smooth, non-myopic trajectories that deviate from the topological edge to maximize information gain. The planner actively directs its FLC (yellow sector) toward high-entropy regions in its belief map $B^C$ (white areas) while concurrently aligning the vehicle to sweep its DLC footprint (blue dashed squares) directly over high-probability coral candidates (cyan circles). Executing in a receding-horizon manner ($N_{exec} = 4$), 
        \texttt{HIMoS} seamlessly incorporates new observations to update the belief map—as evidenced by the newly discovered candidates at $1502$\,s—and replans the subsequent trajectory, successfully balancing broad-area information gathering with biological target sampling.
  
    \subsection{Computational Cost Analysis}
\begin{figure}[tbp]
            \centering
            \includegraphics[width=0.4\textwidth]{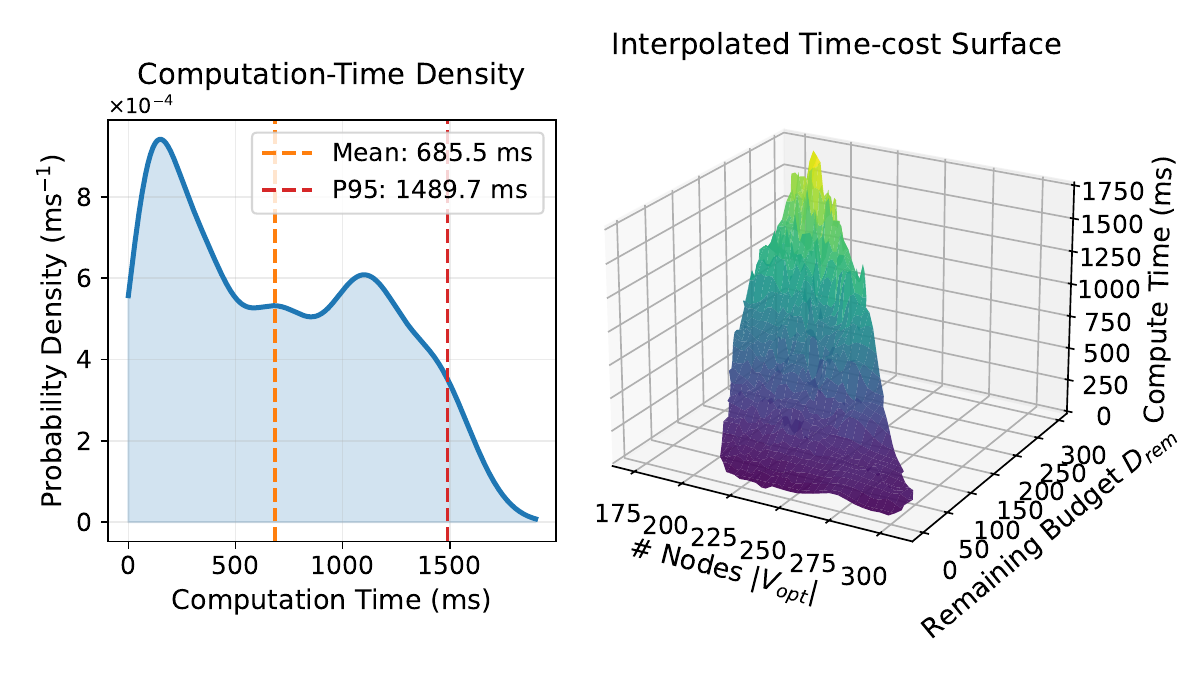}
            \caption{Global Planner computation cost. Left: Solve-time distribution. Right: Time-cost surface w.r.t. number of active nodes $\mathcal{V}_{opt}$ and remaining budget $D_{rem}$.}
            \label{fig:computation_cost}
        \end{figure}

        Real-time performance is critical for embedded AUVs. Although the current implementation is an unoptimized Python prototype, experiments on a Jetson AGX Orin show that the Local Planner solves the NLP in $0.5$ s on average. Executing $N_{exec}=4$ steps provides a $2.0$ s execution horizon, masking computation latency. For the Global Planner, OP runtime scales with active nodes and remaining budget (Fig.~\ref{fig:computation_cost}). Because replanning is event-triggered upon reaching a target region, updates are infrequent. With $95\%$ of global calls finishing within $1.5$ s, \texttt{HIMoS} supports real-time deployment.
                
\section{Conclusion}
    
    We presented \texttt{HIMoS}, a hierarchical multi-modal informative planning framework for fixed-altitude sparse target search. By integrating an adaptive orienteering Global Planner with this local trajectory optimization, \texttt{HIMoS} seamlessly balances wide-area acoustic exploration, visual scouting, and precise target sampling. Our differentiable belief dynamics is a general formulation applicable to any probabilistic sensor model given a spatial FOV.
    Evaluated on real-world benthic datasets with sensor settings reflecting real deployment conditions, the algorithm delivers efficient, robust real-time performance on an embedded AUV computer, indicating strong sim-to-real readiness.
    While the current global-to-local budget conversion employs a coarse linear heuristic, future work will investigate a rigorous, dynamics-aware allocation mechanism and fully deploy \texttt{HIMoS} on a physical AUV for real-world ocean monitoring missions.


\printbibliography{}

\end{document}